\documentclass[letterpaper, 10 pt, conference]{ieeeconf}  

\IEEEoverridecommandlockouts                              

\usepackage{cite}
\usepackage{multicol}
\usepackage{subcaption}
\usepackage{comment}
\usepackage{url}
\usepackage[draft]{hyperref}%
\usepackage{graphicx, epstopdf}
\usepackage{amsmath}
\usepackage{amssymb}
\usepackage{nameref}
\usepackage{color}
\usepackage{comment}
\usepackage{mathtools}

\def\beq{\begin{equation}}
\def\eeq{\end{equation}}
\def\beqa{\begin{eqnarray}}
\def\eeqa{\end{eqnarray}}
\def\ba{\begin{eqnarray*}}
\def\ea{\end{eqnarray*}}
\def\bc{\begin{center}}
\def\ec{\end{center}}
\def\ma{\left[ \begin{array}}
\def\ema{\end{array}\right]}

\def\rr#1{\mathbb{R}^{#1}}

\def\qdbar{{\underline q}_d}

\def\rr#1{\mathbb{R}^{#1}}

\def\calG{{\cal G}}

\def\qbar{{\underline q}}

\title{Neural-Learning Trajectory Tracking Control of Flexible-Joint Robot Manipulators with Unknown Dynamics
}

\author{Shuyang Chen$^{1}$, \textit{Student Member, IEEE} and John T. Wen$^{2}$, \textit{Fellow, IEEE}
\thanks{This research was sponsored 
in part by the New York State Matching Grant program and by the Center for Automation Technologies and Systems (CATS)  at Rensselaer Polytechnic Institute under a block grant from the New York State Empire State Development Division of Science, Technology and
Innovation (NYSTAR).}
\thanks{$^{1}$Shuyang Chen is with the Department of Mechanical, Aerospace, and Nuclear Engineering, Rensselaer Polytechnic Institute, 110 8th St, Troy, NY 12180, USA
        {\tt\small chens26@rpi.edu}}%
\thanks{$^{2}$John T. Wen is with the Department
of Electrical, Computer, and Systems Engineering, Rensselaer Polytechnic Institute, Troy, NY 12180 USA.
        {\tt\small wenj@rpi.edu}}%
}

\begin{document}

\maketitle
\thispagestyle{empty}
\pagestyle{empty}

\begin{abstract}

Fast and precise motion control is important for industrial robots in manufacturing applications. 
However, some collaborative robots sacrifice precision for safety, particular for high motion speed.
The performance degradation is caused by the inability of the joint servo controller to address the uncertain nonlinear dynamics of the robot arm, e.g., due to joint flexibility.
We consider two approaches to improve the trajectory tracking performance through feedforward compensation. %
The first approach uses iterative learning control, with the gradient-based iterative update generated from the robot forward dynamics model.
The second approach uses dynamic inversion to directly compensate for the robot forward dynamics.  If the forward dynamics is strictly proper or is non-minimum-phase (e.g., due to time delays), its stable inverse would be non-causal. 
Both approaches require robot dynamical models. This paper presents results of using recurrent neural networks (RNNs) to approximate these dynamical models -- forward dynamics in the first case, inverse dynamics (possibly non-causal) in the second case.  We use the bi-directional RNN to capture the noncausality. 
The RNNs are trained based on a collection of commanded trajectories and the actual robot responses.
%
We use a Baxter robot to evaluate the two approaches. The Baxter robot exhibits significant joint flexibility due to the series-elastic joint actuators. Both approaches achieve sizable improvement over the uncompensated robot motion, for both random joint trajectories and Cartesian motion.  The inverse dynamics method is particularly attractive as it may be used to more accurately track a user input as in teleoperation.
\end{abstract}


\section{INTRODUCTION}

Motivated by the need for fast and precise motion in manufacturing, robot tracking control has long been of interest \cite{qu1995robust}.  A common joint torque control architecture consists of linear joint position and velocity feedback for stabilization and model-based feedforward for trajectory tracking.  If the feedforward is expressed in the linear-in-parameter form, adaptive control may be applied to estimate the uncertain parameters online while achieving perfect tracking asymptotically \cite{slotine1987adaptive}. 
To further remove the requirement for the expression of the nonlinear robot dynamics, iterative learning control (ILC) \cite{arimoto1990learning} may be used, but only for a specified desired trajectory.  Neural networks (NNs) based controller has been proposed to extend the learning paradigm to arbitrary desired trajectory \cite{Miyamoto1988}.  These approaches have been extended to flexible joint robots, as in some collaborative industrial robots and space robots \cite{brogliato1995global,wang1995simple,lee1998adaptive}.

\begin{figure}[tb]
\centering
\includegraphics[width=0.5\textwidth]{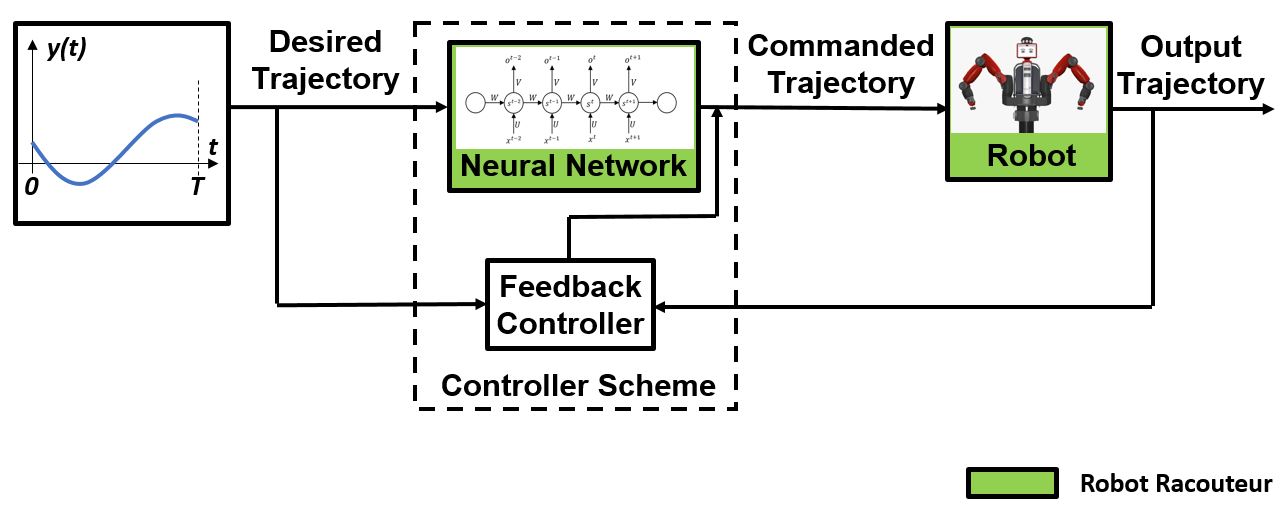}
\caption{Block diagram of the proposed trajectory tracking flow.}
\label{fig:overall_structure}
\end{figure}

Most industrial robots already have a joint torque controller and only allow the joint setpoint adjustment at certain rate.  This paper considers neural-learning control for the outer loop robot tracking control for a flexible joint robot (Baxter) based on recurrent neural networks (RNNs). It has already been demonstrated that RNNs can be used to approximate dynamical systems due to their internal recurrent structure~\cite{Ogunmolu2016}. Here, we propose and compare two approaches based on deep RNNs. Fig.~\ref{fig:overall_structure} shows the schematics  of the control architecture.
The RNN is used as a feedforward controller to produce the control input, which is combined with the feedback term to determine the commanded input to the robot. For the first approach, we train a deep unidirectional RNN to approximate the forward dynamics of the robot manipulator by collecting the response data of the manipulator for a large amount of specified trajectories. Then, we iteratively refine a given desired trajectory offline with the learned model to get the optimal input for the trajectory. Compared to the iterative refinement implemented on the physical robot, our method is much more time efficient. For the second approach, we train another RNN to approximate the inverse dynamics of the robot manipulator. Considering the non-causality of the inverse dynamical system, we train a bidirectional RNN (BRNN)~\cite{Schuster1997BidirectionalRN} with the same set of collected data. We report these two approaches here out of two reasons. First, they can both improve the trajectory tracking performance, given that the entire desired trajectory is known in advance. Second, we have the similar workload for RNNs training due to the one-time data collection. However, for a desired trajectory that is not determined completely in advance but is generated online via user teleoperation, the dynamical inversion approach is preferred for online dynamics compensation with its capability of real-time inference. The first approach, which requires the knowledge of the entire trajectory for iterative refinement, is not suitable anymore. To our best knowledge, there are few works exploring and comparing the two approaches using RNN/BRNN as feedforward compensation for robot manipulators trajectory tracking control with unknown dynamics.

To validate the effectiveness and generalization capability of the proposed approaches. We conduct experiments for tracking an unseen multi-joint sinusoidal trajectory, a random joint trajectory, and a Cartesian trajectory using two approaches. For those entirely given desired trajectories, the corresponding feedforward controller input trajectory is obtained either through ILC with the trained RNN (the first approach), or directly filtered through the BRNN (the second approach). We test that, for either of the two approaches, we can apply additional iterations of ILC with the resulted feedforward trajectory on the physical manipulator to further improve the tracking performance. We also validate the effectiveness of the dynamical inversion approach for real-time online compensation of a user-teleoperated trajectory. Instead of obtaining the entire feedforward input trajectory offline, the feedforward input at each time step is obtained by the trained BRNN in real-time. For all above experiments, at each time step, the feedforward control input is combined with a feedback term obtained by a proportional feedback controller, and the combined input will be used as the commanded joint setpoint to the robot manipulator. By comparing with the performance of a baseline feedback controller, we demonstrate that both of two approaches work effectively in reducing the trajectory tracking error.

The rest of the paper is organized as follows. Section~\ref{Related Work} summarizes related work of advanced approaches on trajectory tracking control, including NNs, ILC, and differential dynamic programming (DDP), and highlights the difference from our approaches. Section~\ref{problem statement} states the problem, followed by the detailed description of methodology in Section~\ref{methodology}. The experimental results are presented in Section~\ref{results}. Finally, Section~\ref{conclusions and future work} concludes the paper and adds some new insight.

\section{Related Work}
\label{Related Work}

There have been a lot of works for improving trajectory tracking performance of robot manipulators through controllers design based on learning dynamics models or inverse dynamics through NNs~\cite{Jiang2017}. NNs, with their strong generalization and approximation capability given enough training data, have attracted more and more attention in control community for controller design. A detailed review of neural-learning control can be referred in~\cite{JIN201823}.

Candidate NNs, including radial basis function (RBF) NN, RNN, and multi-layer feedforward NN, have been explored to approximate the robot forward dynamics model for adaptive
trajectory tracking controller design~\cite{Yoo2008, Perez-Cruz2014, He2016, Wang2017NeuralL, He2018NNLS}. In~\cite{Yang2017}, an adaptive controller designed using RBF NN producing joint torque input was proposed to compensate the unknown dynamics and a payload for a Baxter robot. Compared to their work, instead of using torque level control, we use joint setpoint control. 
Also, it is a non-trivial task to determine the number of Gaussian kernels and their centers as well as shapes for RBF NN. They trained a large amount of parameters (in the order of 
$10^5$) for RBF NN to approximate the robot dynamics model, whereas we have much less amount of parameters (in the order of $10^3$) considering that the parameters are shared among all RNN time step cells. In addition, in their work, Lyapunov stability theory was applied to guarantee the bounded tracking error and also derive the NNs parameters update law. However, there is no guarantee that the model estimation error will also converge. They used separate RBF NNs to approximate robot dynamical model components individually including inertial matrix $M$, Coriolis matrix $C$, and gravity torque $G$ whereas they did not guarantee the properties of the robot dynamics model such as the positive-definiteness of $M$. On the contrary, we use supervised learning to train the RNNs given collected input/response data to directly approximate the robot internal dynamics as one integrated model. Finally, the dynamics model they tried to approximate is for rigid-joint robot manipulators instead of flexible-joint ones.

Works have also been reported of using NNs to learn an inverse dynamics model for dynamics compensation~\cite{Talebi1998}. Li et al.~\cite{Li2017} proposed an NN-based control architecture with an NN pre-cascaded to a quadrotor under a feedback controller. 
Compared to their approach, our method is different in the following aspects. First of all, we apply the trained RNNs to an articulated robot manipulator, which has very different dynamics from the quadrotors. Next, we train RNNs to approximate the feedfoward and inverse dynamics while they use multi-layer feedforward NNs. And specifically we train a BRNN to address the noncausality of the inverse of a strictly proper system.
It has been also shown that RNN can outperform Gaussian Processes~\cite{Calandra2015} in performance of learning the inverse dynamics of a robot manipulator with linear time complexity, given sufficient training data~\cite{Rueckert2017}.

\begin{figure}[tb]
\centering
\includegraphics[width=0.47\textwidth]{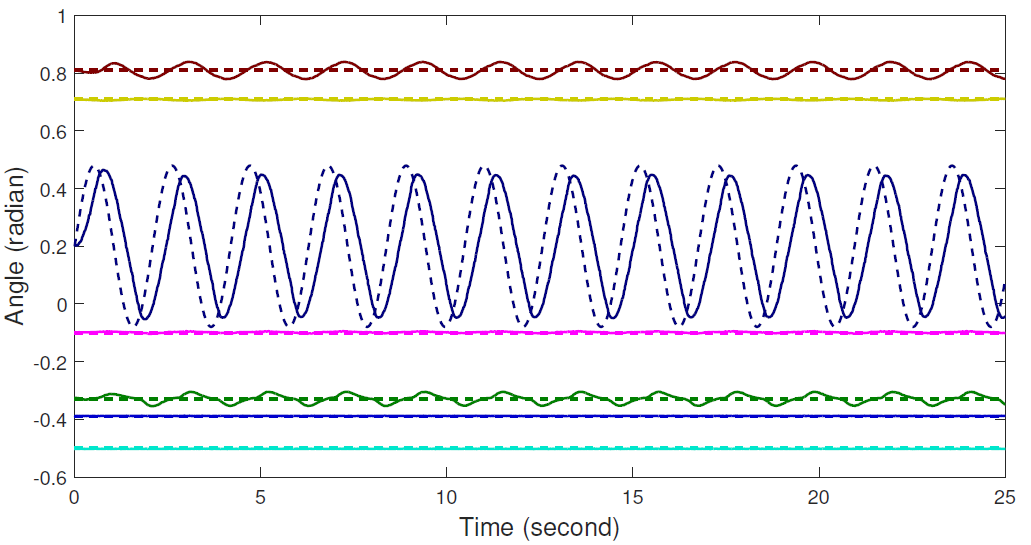}
\caption{Proof of coupling of Baxter joints. We only command active motion of joint 4 (the blue sinusoid in the middle of the figure) with other joints fixed. The dashed and solid lines indicate desired and actual joint angles, respectively. The lines of different colors indicate different joints. The figure shows that the output of some joints have similar motion pattern to joint 4, although they are commanded not moving. }
\label{fig:coupling_test} 
\end{figure}

ILC is another approach to improve the tracking performance by searching for the optimal input trajectory for a desired trajectory via iteratively updating the control input with previous tracking errors. The main drawback of ILC is its non-transferability. For each new desired trajectory, ILC must be re-implemented to search for the optimal input. 
In our work~\cite{chen2019}, we trained multi-layer feedforward NNs offline to find a good estimate of the inverse dynamics of an industrial ABB robot for trajectory tracking control, by implementing ILC for a large number of desired trajectories to collect training data in a high-fidelity simulator. We also applied transfer learning to transfer the learned representations in simulator to the real-world by fine-tuning the NNs with real-world data. However, the above techniques cannot be applied to Baxter robots due to the following reasons. First, there is no good dynamic simulator for Baxter, which makes it tedious to implement ILC on the physical robot for a large amount of trajectories. 
Then, the joints of ABB robot are decoupled, thus we can train 6 NNs, with each NN approximating the dynamical inversion for one joint. But the joints of Baxter are coupled as shown in Fig.~\ref{fig:coupling_test}. Thus, we must train one NN to couple all 7 joints instead of training separate NNs. To address this, we train RNNs with 2D inputs, which are composed of roll-out of a segment of a joint trajectory for all 7 joints. Finally, since we directly collect training data from the physical robot, no transfer learning is needed.

DDP~\cite{ddp} is another gradient-based algorithm that uses a locally-quadratic dynamics model to search for local optimal control inputs. For nonlinear systems, the control inputs will be iteratively updated by applying the linear quadratic regulator (LQR) to the linearized system about the current trajectory. In general, it is impractical to solve DDP at run-time.

\section{Problem Statement}
\label{problem statement}
Here we aim at achieving improved trajectory tracking for the left arm of a Baxter robot. The arm is under closed loop joint servo control with input $u\in\rr n$ (here we use the joint position command $q_c$), and the output $q\in\rr n$, the measured joint position. Our goal is to compensate the robot inner loop dynamics $\calG$, by learning a mapping from the desired trajectory $q_{d}$ to the feedforward control $q_f$ (as shown in Fig.~\ref{fig:control_structure_2}), which will be combined with real-time feedback control input to produce the commanded input $q_c$ to achieve improved tracking performance of the overall system, compared to the baseline feedback controller.

\section{Methodology}
\label{methodology}

\subsection{Proposed Control Law}

\begin{figure}[tb]
\centering
\includegraphics[width=0.45\textwidth]{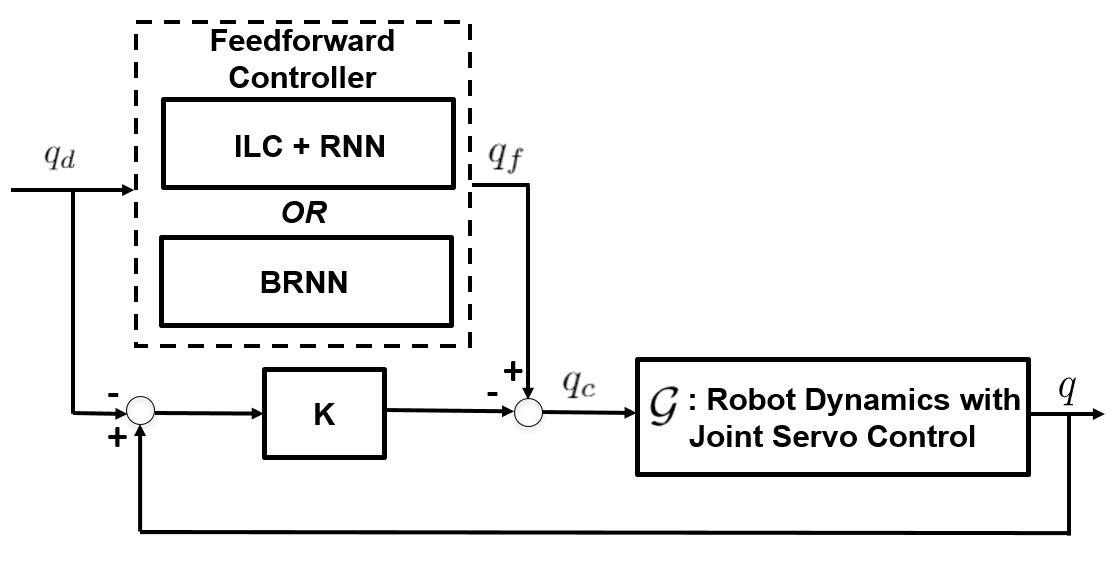}
\caption{Overall control architecture.}
\label{fig:control_structure_2} 
\end{figure}

We propose the controller composed of feedforward and feedback components indicated in Fig.~\ref{fig:control_structure_2}. The associated control law is as below:

\begin{equation}
q_c(t+1) = q_f(t+1)-k\cdot (q(t)-q_{d}(t))
\label{eq:ff+fb control law}
\end{equation}

where $q_{d}(t)$ is the desired joint position at time $t$, $q(t)$ is the real time output of robot, and $q_f(t+1)$ is the feedforward command at time $t+1$, which is obtained either by implementing ILC with the trained unidirectional RNN that approximates the manipulator forward dynamics, or directly filtering by the dynamical inversion approximated by the BRNN. $q_c(t+1)$ is the resulting commanded joint input for robot manipulator at time $t+1$.

On the other hand, the stable baseline feedback controller is:

\begin{equation}
q_c(t+1) = q_d(t+1)-k\cdot (q(t)-q_{d}(t))
\label{eq:fb control law}
\end{equation}

\subsection{Feedforward Controller using RNN and ILC}

The feedforward control input $q_f$ can be obtained by implementing ILC with the trained RNN. In order to train an RNN with large capacity of memory by supervised learning, we need a large amount of input/output training data.

\subsubsection{Data Collection}
\begin{figure}[tb]
\centering
\includegraphics[width=0.5\textwidth]{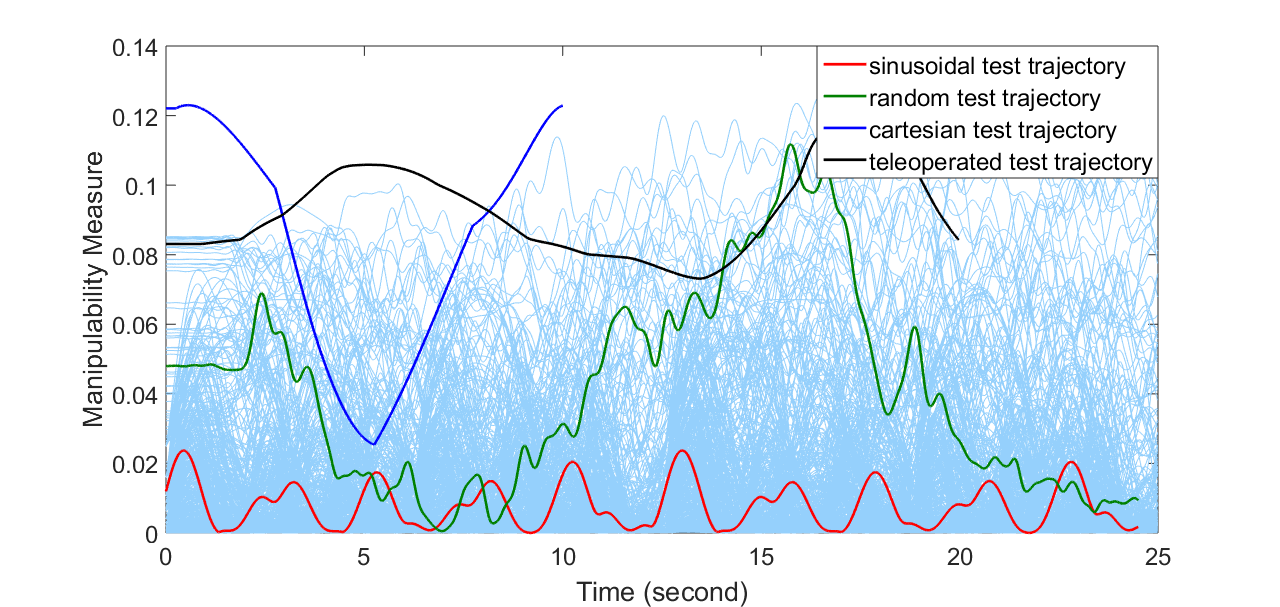}
\caption{Manipulability measure over 500 training joint trajectories (shown as the light blue background) and 4 testing joint trajectories.}
\label{fig:manipulability_measure} 
\end{figure}

We collect raw response joint position data of the left arm of Baxter for in total 500 joint trajectories $q_d$, among which 100 trajectories are purely random and another 400 are sinusoids with varying magnitudes and frequencies. We choose the trajectories by obeying the joint limits of Baxter, as well as trying to cover a feasible portion of Baxter joint space. The manipulability measure~\cite{Yoshikawa1985} of 500 training joint trajectories as well as the 4 testing joint trajectories (described in Section~\ref{results}) are plotted in Fig.~\ref{fig:manipulability_measure}.
Each trajectory $q_d$ contains 2500 joint setpoints for all 7 joints (thus a 7 by 2500 matrix) and is commanded to Baxter at 100 Hz. At the same time, the output joint trajectory $q$ of the manipulator is collected. Then, our goal is to obtain an approximation of the forward dynamics $\calG$ by learning from a set of ($q_d$, $q$).

\subsubsection{Unidirectional RNN Training}

\begin{figure}[tb]
   \centering
   \begin{subfigure}[b]{0.49\textwidth}
        \includegraphics[width=\textwidth]{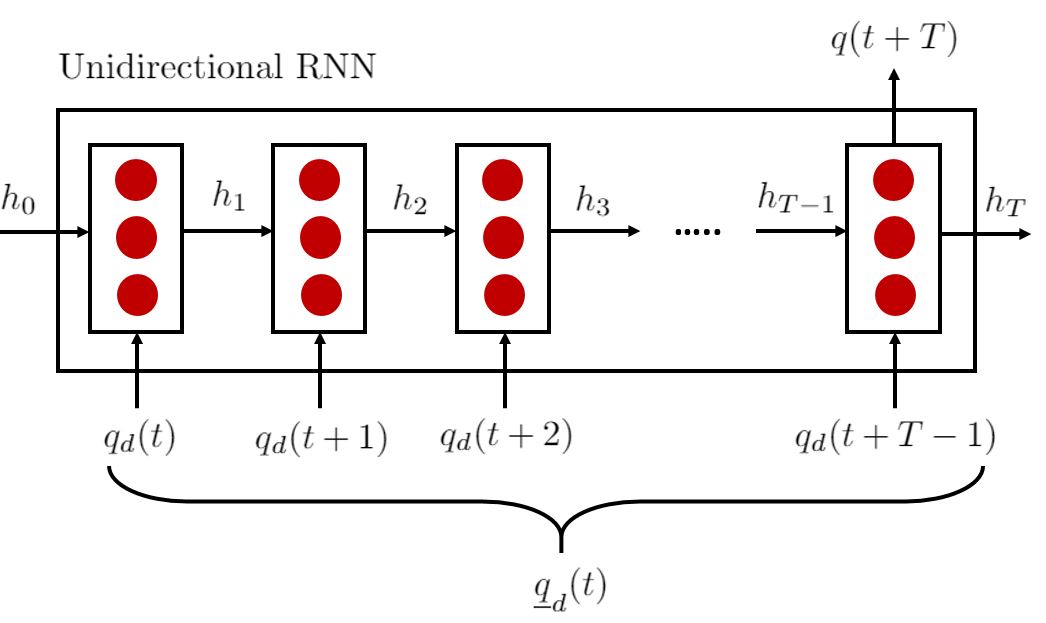}
        \caption{Unrolled unidirectional RNN with $T$ time steps of inputs.}
        \label{fig:unrolled_unidirectional_RNN}
    \end{subfigure}
    \hfill
    \begin{subfigure}[b]{0.49\textwidth}
        \includegraphics[width=\textwidth]{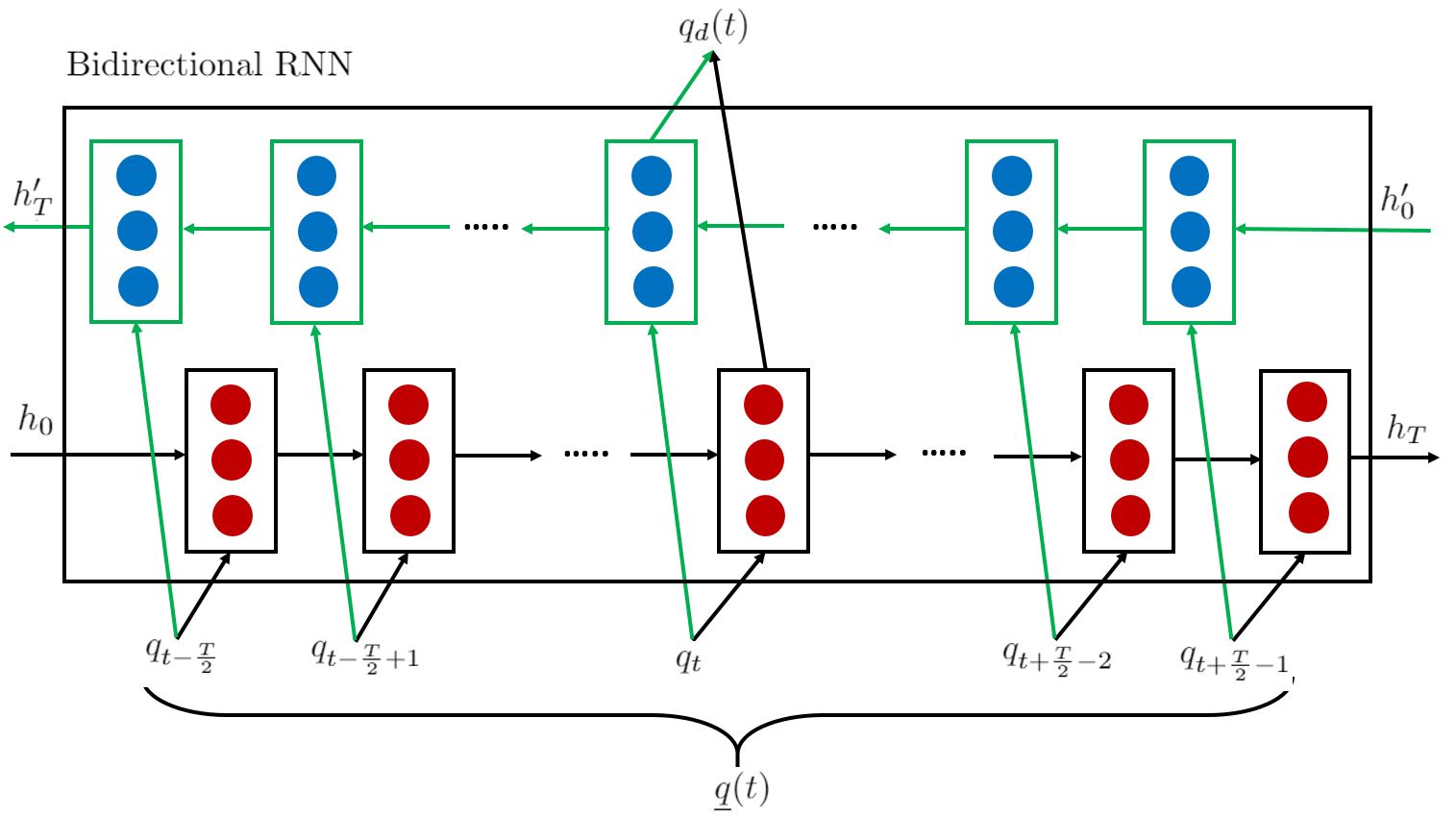}
        \caption{Unrolled bidirectional RNN with $T$ time steps of inputs.}
        \label{fig:unrolled_bidirectional_RNN}
    \end{subfigure}
   \caption{Unrolled RNNs structures. $h_i$ represents the internal states of each time step.}
   \label{fig:unrolled RNNs}
   \vspace{-5pt}
\end{figure}

We train a unidirectional RNN as shown in Fig.~\ref{fig:unrolled_unidirectional_RNN} to represent the forward dynamics of the manipulator considering the causality of the dynamical system, and the coupling of Baxter joints.

In specific, during training, the input to the RNN is a sequence of a desired joint trajectory containing $T$ time steps setpoints $\qdbar(t):=\{q_d(\tau): \tau\in[t,t+T-1]\}$. Each $q_d(\tau)$ is a 7 by 1 vector denoting the joint position setpoint on a desired trajectory $q_d$ for all 7 joints at time $\tau$. The output of the RNN is the joint position $q(t+T)$ (also a 7 by 1 vector) of the output trajectory $q$. Thus, for each input $q_d$ and output $q$ trajectory containing 2500 joint position setpoints, we can get $2500-T$ training samples $(\qdbar(t), q(t+T))$.

We choose $T=50$ to guarantee that the input of RNN contains enough dynamics information for the RNN to model. In total we have 1225000 samples, and we use 80~\% of the samples for training and 20~\% for testing. We train a 4-layer unidirectional RNN with Gated Recurrent Units (GRUs) instead of LSTM cells after comparison in \texttt{TensorFlow}. We use \texttt{AdamOptimizer} with an initial learning rate of $1\times 10^{-3}$ to tune the parameters of RNN by minimizing the mean squared error (MSE) between the predicted output of RNN and the expected output over a randomly selected batch of 256 training samples in each training iteration. Also, we use dropout~\cite{JMLR:v15:srivastava14a} of 0.5 to avoid overfitting. The training process converges after 10000 training iterations.

After training, the RNN can provide a mapping from a sequence of 50 time steps ($T=50$) joint position inputs (a 7 by 50 matrix) $\qdbar(t)$ to the predicted output $q_o(t+50)$ (a 7 by 1 vector):

\begin{equation}
    \qdbar(t):=\{q_d(\tau): \tau\in[t,t+49]\} \rightarrow q_o(t+50)
\end{equation}

\subsubsection{Iterative Learning Control}
We can implement ILC with the trained RNN that emulates the forward dynamics of the manipulator to search for the optimal input for a given desired trajectory. We use the multiple-input multiple-output (MIMO) gradient-based ILC algorithm as developed in~\cite{chen2019} based on the work in~\cite{Potsaid2007}. The key idea of the algorithm is that, we iteratively update the input $u^{k}$ ($k$ represents the number of iterations) until convergence with the gradient of the tracking error with respect to the current input $u^{k}$ as below:

\begin{equation}
u^{k+1} = u^{k}-\alpha_k G^*(s) e_q
\label{eq:ILC}
\end{equation}

where $G^*(s)$ represents the adjoint of a linear time invariant system $G(s)$ approximating the internal dynamics $\calG$, and $e_q:=(q^k-q_d)$, which is the tracking error using $u^k$ as input. $G^*(s) e_q$ represents the gradient of the tracking error with respect to $u^{k}$. $\alpha_k$ is the optimal learning rate for $k^{th}$ iteration and can be obtained by a 1-d line search. The above iterative refinement algorithm is easily implemented with the trained RNN. $u^0$ is simply chosen to be the desired output trajectory $q_d$ to start the iteration. 

The drawback of the above method is that although it is much faster (compared to implementation on the physical system) to implement ILC with the RNN offline to find the feedforward control input for a given desired trajectory, it still takes several seconds until the convergence of ILC. Thus, it is only suitable for an entirely given trajectory and will not work for real-time compensation.

\subsection{Feedforward Controller using BRNN}

The feedforward control input $q_f$ can also be obtained by directly filtering $q_{d}$ through the trained BRNN, which is used to approximate the manipulator inverse dynamics. In this way, there are no iterations required thus it is suitable for real-time compensation.

\subsubsection{Data Collection} 
Instead of collecting training data through ILC as we did in~\cite{chen2019}, we can use the same collected training data above but use the output trajectory $q$ as input and the desired trajectory $q_d$ as output to the BRNN, inspired by~\cite{Li2017}. The idea behind this is that, if $q_d$ and $q$ are corresponding input and output trajectories for a dynamical system $\calG$, then with input of $q$, the inverse dynamical system would produce output of $q_d$.

\subsubsection{Bidirectional RNN Training} 

Here, we train a BRNN as shown in Fig.~\ref{fig:unrolled_bidirectional_RNN} to represent the inverse dynamics of the manipulator considering the possible noncausality of the inverse of a non-minimum-phase system.

In specific, the input to the BRNN during training is a sequence of an output joint trajectory $q$ containing $T$ time steps $\qbar(t):=\{q(\tau): \tau\in[t-T/2,t+T/2-1]\}$. The output of the BRNN is the input joint position $q_d(t)$. Thus, for each input $q_d$ and output trajectory $q$ containing 2500 joint position setpoints, we can get 2501-$T$ training samples $(\qbar(t), q_d(t))$.

Similar to the training of the unidirectional RNN, we choose $T=50$, which will result in 1225500 samples, and we use 80~\% for training and 20~\% for testing. We train a 2-layer BRNN with GRU cells in \texttt{TensorFlow} by \texttt{AdamOptimizer} with an initial learning rate of $1\times 10^{-3}$ to tune the parameters by minimizing the MSE between the predicted output of BRNN and the expected output over a randomly selected batch of 256 training samples in each training iteration.

After training, the BRNN can provide a mapping from a sequence of 50 time steps of desired joint position inputs (a 7 by 50 matrix) $\qdbar(t)$ to a feedforward control input $q_f(t)$ (7 by 1):

\begin{equation}
    \qdbar(t):=\{q_d(\tau): \tau\in[t-25,t+24]\} \rightarrow q_f(t)
    \label{brnn}
\end{equation}

\subsection{Resolved Velocity Controller for Human Teleoperation}

To facilitate the human teleoperation of the redundant manipulator, we develop the resolved velocity controller based on our work in~\cite{chen2018} as below:

\begin{equation}
\min_{\dot q, \alpha_r, \alpha_p}{||J\dot q - \alpha v_d||^2 + \epsilon_r(\alpha_r - 1)^2 + \epsilon_p(\alpha_p - 1)^2} 
\label{eq:qp}
\end{equation}
subject to equality constraints such as orientation control: $h_E(q) = 0$, and
inequality constraints such as joint limits: $h_I(q)>0$,
where $J$ is the Baxter left arm Jacobian and $\dot q$ is the resolved robot joint velocity for the user commanded velocity $v_d$. And 
$$
\alpha =
\begin{bmatrix}
\begin{array}{ccc}
\alpha_r I_{3\times 3} & 0_{3\times 3} \\
0_{3\times 3} & \alpha_p I_{3\times 3} \\
\end{array}
\end{bmatrix}
$$
in which $\alpha_r$ scales the angular velocity part of the human commanded velocity $v_d$, and $\alpha_p$ 
scales the linear velocity part. $\epsilon_r, \epsilon_p$ are two scaling factors. The resulted trajectory will be used as input to the BRNN to produce the feedforward control input $q_f$ in real-time.

\subsection{Robot Raconteur Bridge}

We use Robot Raconteur (RR)~\cite{Wason2011} as middleware to coordinate robot manipulator control and communication, joystick teleoperation, and feedforward controller input inference using the trained RNNs with separate RR services, as shown by the green blocks in Fig.~\ref{fig:overall_structure}. The overall coordination is conducted by a MATLAB script that connects to these services as a client. The RR-Baxter bridge software is available at \htmladdnormallink{https://github.com/rpiRobotics/baxter\_rr\_bridge}{https://github.com/rpiRobotics/baxter\_rr\_bridge}.

\section{Results}
\label{results}

\subsection{Tracking a Multi-Joint Trajectory}
\begin{figure*}[tb]
\centering
\begin{multicols}{2}
\hspace*{0.3in}
\includegraphics[width=0.9\textwidth, height=0.45\textwidth]{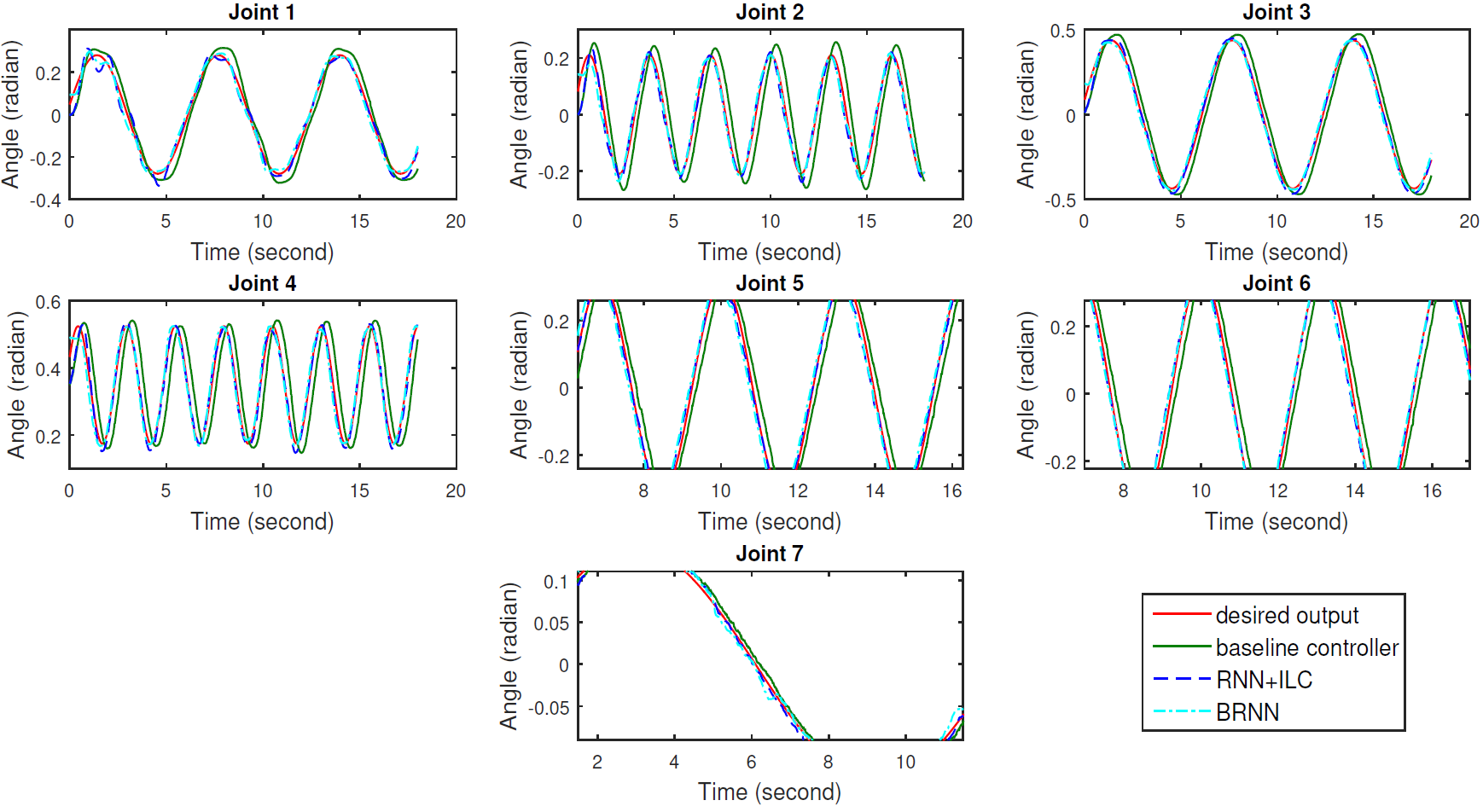}
\end{multicols}
\caption{Comparison of tracking performance without and with
the RNN/BRNN feedforward controllers of sinusoidal joint trajectories. The plots of joint 5 to 7 are zoomed in to better visualize the difference.}
\label{fig:track_sin_plot_two_approaches} 
\end{figure*}

Fig.~\ref{fig:track_sin_plot_two_approaches} 
shows the results of tracking an unseen sinusoidal joint trajectory using two approaches. Table~\ref{error_7_joint_path} compares the tracking errors in terms of $\ell_2$ and $\ell_\infty$ norms (using the error vectors at each sampling instant) for each joint, using the baseline feedback controller, the first approach and the second approach, respectively. From the figure and the table, we can see that the feedforward control input obtained by either of the two approaches can improve the tracking performance a lot compared to the baseline feedback controller (in average over 50~\% of improvement). Also, the second approach works slightly better than the first method. One possible reason is the initial transient process at the beginning when using ILC~\cite{chen2019}.

\begin{table*}[tb]
\caption{$\ell_2$ and $\ell_\infty$ norm of tracking errors of 7-dimensional sinusoidal joint trajectories using baseline controller, the first approach and the second approach.}
\label{error_7_joint_path}
\begin{center}
\begin{multicols}{2}
\begin{tabular}{|c|c|c|c|}
\hline
Joint & Baseline Controller Tracking Error (rad) & RNN + ILC + Feedback Tracking Error (rad) & BRNN + Feedback Tracking Error (rad) \\
 & $\ell_2 \hspace{1.5cm} \ell_\infty$ & $\ell_2 \hspace{1.5cm} \ell_\infty$
 & $\ell_2 \hspace{1.5cm} \ell_\infty$ \\
\hline
1 & 2.3816 \hspace{1cm} 0.0973 & 1.3622 \hspace{1cm} 0.0884 & 0.8088 \hspace{1cm} 0.0590\\
2 & 4.0453 \hspace{1cm} 0.1296 & 1.2072 \hspace{1cm} 0.1124 & 0.7163 \hspace{1cm} 0.0545  \\
3 & 3.4910 \hspace{1cm} 0.1029 & 1.2245 \hspace{1cm} 0.0942 & 0.7328 \hspace{1cm} 0.0727 \\
4 & 4.1703 \hspace{1cm} 0.1437 & 0.9874 \hspace{1cm} 0.0962 & 0.5944 \hspace{1cm} 0.0391 \\
5 & 2.5834 \hspace{1cm} 0.0917 & 1.0075 \hspace{1cm} 0.0672 & 0.7247 \hspace{1cm} 0.0530 \\
6 & 2.9346 \hspace{1cm} 0.0955 & 1.1818 \hspace{1cm} 0.0808 & 0.7121 \hspace{1cm} 0.0593 \\
7 & 0.3359 \hspace{1cm} 0.0133 & 0.2845 \hspace{1cm}  0.0178 & 0.2523 \hspace{1cm} 0.0155 \\
\hline
\end{tabular}
\end{multicols}
\end{center}
\end{table*}

\subsection{Tracking a Random Joint Trajectory}
We did another test of tracking an unseen random joint trajectory using two approaches. Table~\ref{error_7_random_path} summarizes the results of tracking errors. From the table, we can see that the feedforward control input obtained by both approaches can improve the tracking performance obviously compared to the baseline feedback controller (in average over 40~\% of improvement). Also, the second approach still works slightly better than the first one.

\begin{table*}[tb]
\caption{$\ell_2$ and $\ell_\infty$ norm of tracking errors of 7-dimensional random joint trajectories using baseline controller, the first approach and the second approach.}
\label{error_7_random_path}
\begin{center}
\begin{multicols}{2}
\begin{tabular}{|c|c|c|c|}
\hline
Joint & Baseline Controller Tracking Error (rad) & RNN + ILC + Feedback Tracking Error (rad) & BRNN + Feedback Tracking Error (rad) \\
 & $\ell_2 \hspace{1.5cm} \ell_\infty$ & $\ell_2 \hspace{1.5cm} \ell_\infty$
 & $\ell_2 \hspace{1.5cm} \ell_\infty$ \\
\hline
1 & 4.1746 \hspace{1cm} 0.2515 & 2.1068 \hspace{1cm} 0.1495 & 2.1974 \hspace{1cm} 0.1478\\
2 & 3.8348 \hspace{1cm} 0.2129 & 1.9292 \hspace{1cm} 0.1131 & 1.7717 \hspace{1cm} 0.1357  \\
3 & 3.7421 \hspace{1cm} 0.2193 & 2.1502 \hspace{1cm} 0.1642 & 1.9836 \hspace{1cm} 0.1458 \\
4 & 2.9880 \hspace{1cm} 0.1614 & 1.9791 \hspace{1cm} 0.1366 & 1.5006 \hspace{1cm} 0.1015 \\
5 & 1.9861 \hspace{1cm} 0.1156 & 1.7605 \hspace{1cm} 0.0974 & 1.3623 \hspace{1cm} 0.1179 \\
6 & 2.2318 \hspace{1cm} 0.167 & 1.7216 \hspace{1cm} 0.1153 & 1.3887 \hspace{1cm} 0.1223 \\
7 & 1.8733 \hspace{1cm} 0.1448 & 1.3336 \hspace{1cm} 0.1104 & 1.2803 \hspace{1cm} 0.0948 \\
\hline
\end{tabular}
\end{multicols}
\end{center}
\end{table*}

\subsection{Tracking a Cartesian Trajectory}

\begin{figure}[tb]
\centering
\includegraphics[width=0.45\textwidth]{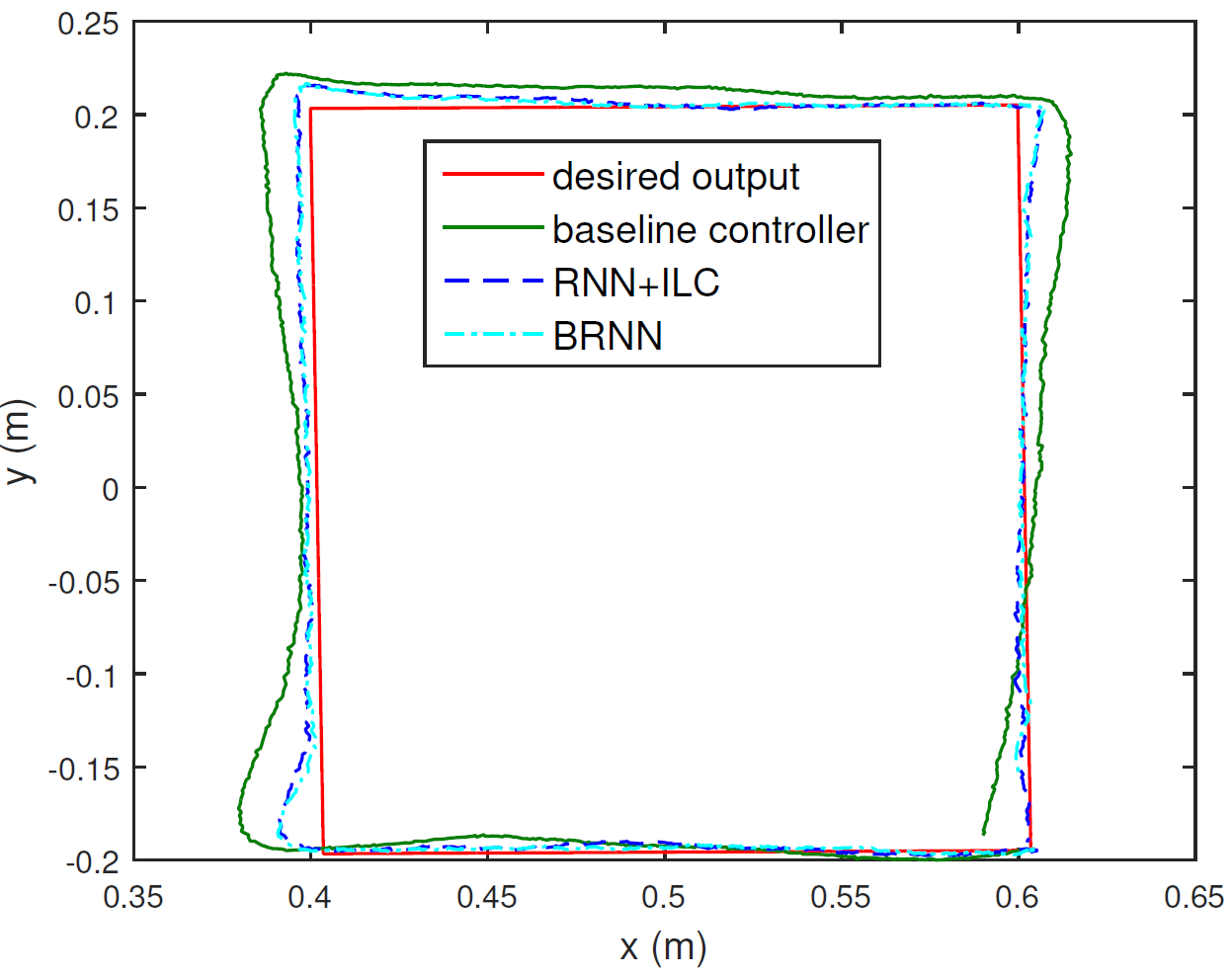}
\caption{Comparison of tracking  performance without and with the RNN/BRNN feedforward controllers for a Cartesian square trajectory in $x-y$ plane with $z$ constant.}
\label{fig:track_cart_plot_xyz} 
\end{figure}

Fig.~\ref{fig:track_cart_plot_xyz} shows the results of using two approaches to track a Cartesian trajectory, which is a square in the $x-y$ plane with $z = 0.2~m$. During the motion, the orientation of the end-effector is fixed. Table~\ref{error_cart_path} compares the corresponding tracking errors for each Cartesian direction. From the figure and the table, we show that the tracking performance is significantly improved with the feedforward control inputs (in average over 50~\%), as compared to the baseline feedback controller. 

\begin{table*}[tb]
\caption{$\ell_2$ and $\ell_\infty$ norm of tracking errors of a Cartesian trajectory in three axes using baseline controller, the first approach and the second approach.}
\label{error_cart_path}
\begin{center}
\begin{multicols}{2}
\begin{tabular}{|c|c|c|c|}
\hline
Axis & Baseline Controller Tracking Error (m) & RNN + ILC + Feedback Tracking Error (m) & BRNN + Feedback Tracking Error (m) \\
 & $\ell_2 \hspace{1.5cm} \ell_\infty$ & $\ell_2 \hspace{1.5cm} \ell_\infty$
 & $\ell_2 \hspace{1.5cm} \ell_\infty$ \\
\hline
$x$ & 0.4478 \hspace{1cm} 0.0363 & 0.1306 \hspace{1cm} 0.0124 & 0.1223 \hspace{1cm} 0.0129\\
$y$ & 0.5694 \hspace{1cm} 0.0459 & 0.1808 \hspace{1cm} 0.0199 & 0.1685 \hspace{1cm} 0.0200  \\
$z$ & 0.0582 \hspace{1cm} 0.0084 & 0.0455 \hspace{1cm} 0.0068 & 0.0273 \hspace{1cm} 0.0023 \\
\hline
\end{tabular}
\end{multicols}
\end{center}
\end{table*}

We also find that the tracking performance for all test cases above can be further improved by implementing additional ILC iterations on the physical robot, using the feedforward-compensated trajectory as the initial input $u^0$ in (\ref{eq:ILC}).

\subsection{Tracking with Human-Teleoperated Motion}

\begin{figure*}[tb]
\centering
\begin{multicols}{2}
\hspace*{0.25in}
\includegraphics[width=0.9\textwidth, height=0.45\textwidth]{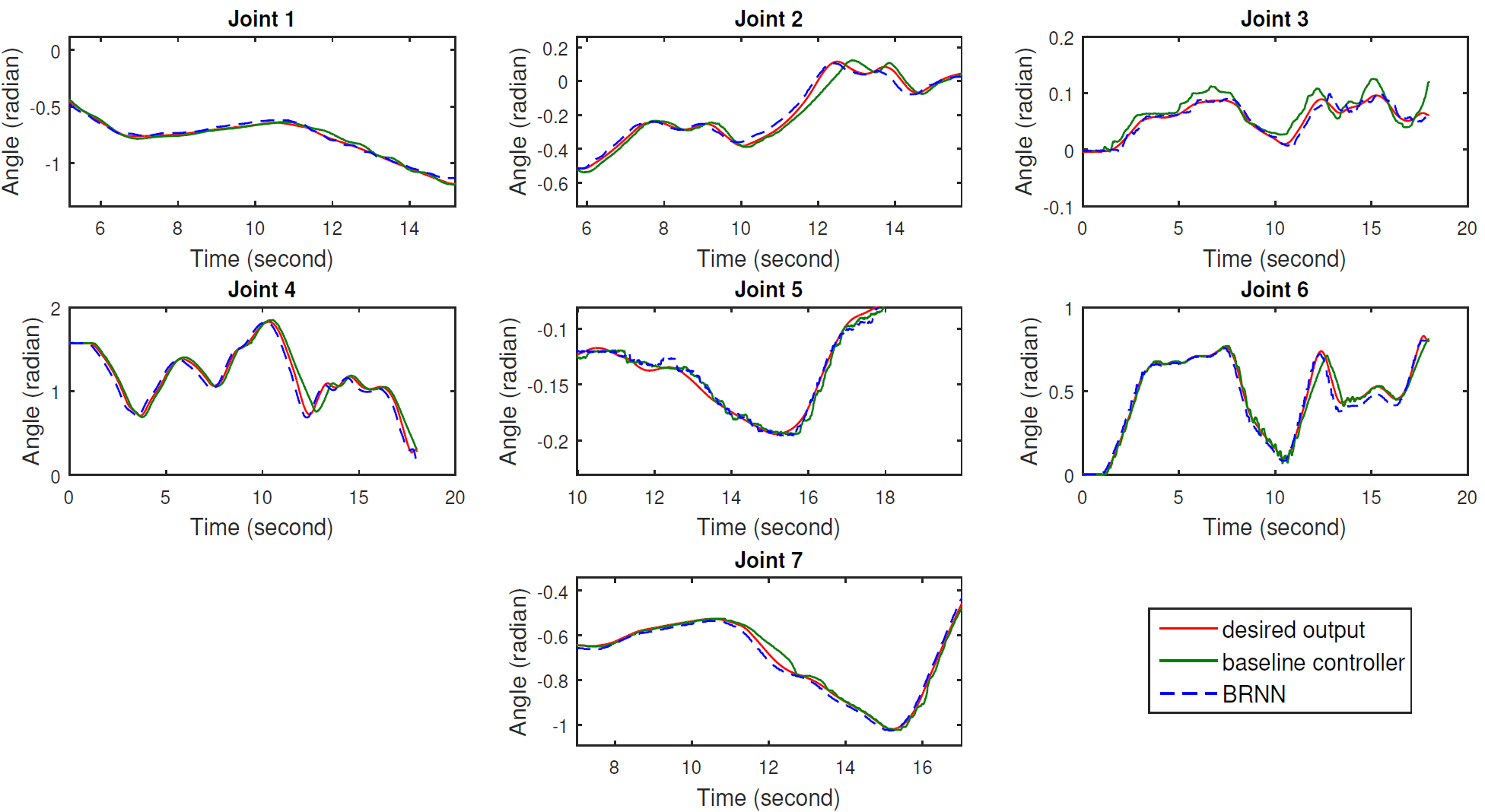}
\end{multicols}
\caption{Comparison of tracking performance without and with
the BRNN feedforward controller of a user teleoperated random joint trajectories. The plots of joint 1, 2, 5 and 7 are zoomed in to better visualize the difference.}
\label{fig:track_teleop_plot} 
\end{figure*}

We may also encounter the case that the desired trajectory is not given completely and the feedforward controller must work in real-time to produce the control input. In this case we can only use the BRNN. To verify the feasibility of the BRNN for online compensation, we use an Xbox joystick to teleoperate the left arm of Baxter using the resolved velocity controller in (\ref{eq:qp}), in which case the desired trajectory is randomly planned online by a user. Note that at each time step $t$, we need to collect future 24 user inputs (as from (\ref{brnn})) to get the required 50 ($T=50$) input joint setpoints of the BRNN, which will cause about 0.24~s delay corresponding to 100 Hz communication rate. The delay can be reduced by using a smaller $T$, but if $T$ is too small, there will not be enough information for the BRNN to capture the inverse dynamics of the manipulator. Fig.~\ref{fig:track_teleop_plot} shows the results of tracking one random teleoperated joint trajectories using the BRNN feedforward controller. Table~\ref{error_tele_path} compares the tracking errors. The figure and the table highlight that the feedforward control input obtained by BRNN can effectively improve the tracking performance for a randomly generated trajectory compared to the baseline feedback controller (in average over 30~\% of improvement). This real-time compensation feature makes it suitable for tracking a trajectory in a timely manner.

\begin{table*}[tb]
\caption{$\ell_2$ and $\ell_\infty$ norm of tracking errors of 7-dimensional user teleoperated joint trajectories using baseline controller and the second approach.}
\label{error_tele_path}
\begin{center}
\begin{tabular}{|c|c|c|}
\hline
Joint & Baseline Controller Tracking Error (rad) & BRNN + Feedback Tracking Error (rad) \\
 & $\ell_2 \hspace{1.5cm} \ell_\infty$ & $\ell_2 \hspace{1.5cm} \ell_\infty$
  \\
\hline
1 & 1.4939 \hspace{1cm} 0.1826 & 1.2083 \hspace{1cm} 0.0954 \\
2 & 3.0560 \hspace{1cm} 0.3547 & 1.2055 \hspace{1cm} 0.0797  \\
3 & 0.6279 \hspace{1cm} 0.0196 & 0.3023 \hspace{1cm} 0.0599 \\
4 & 5.4097 \hspace{1cm} 0.6381 & 2.5719 \hspace{1cm} 0.1529 \\
5 & 0.2238 \hspace{1cm} 0.0265 & 0.1946 \hspace{1cm} 0.0140  \\
6 & 1.4105 \hspace{1cm} 0.1241 & 1.3121 \hspace{1cm} 0.0932 \\
7 & 0.7065 \hspace{1cm} 0.0592 & 0.6404 \hspace{1cm} 0.0344 \\
\hline
\end{tabular}
\end{center}
\end{table*}


\section{Conclusions and Future Work}
\label{conclusions and future work}

In this paper, we explored two approaches for trajectory tracking control of a flexible-joint robot manipulators with unknown dynamics through feedforward compensation by recurrent neural networks. 
The first approach is to implement iterative learning control with the trained RNN which approximates the forward dynamics of the robot for a specified desired trajectory offline. The second one is to filter the desired trajectory by the BRNN which emulates the dynamical system inversion. We first demonstrated the results of two approaches for the manipulator trajectory tracking control when a desired trajectory is entirely given in advance. We also showed the results of tracking a human teleoperated random joint trajectory which is generated online using the BRNN feedforward controller producing real-time compensation control input. By comparing with a baseline feedback controller, the effectiveness of our approaches is verified.

Future work includes the development of an adaptive controller with generalized basis network that can approximate multiple dynamical systems via online parameters tuning. Also, it is worthwhile to explore the transferability of the trained model to another Baxter robot. The tracking performance may be further improved by combining the feedforward control inputs produced by RNNs with more sophisticated feedback control laws instead of a pure proportional control as described here.


\bibliographystyle{IEEEtran}
\bibliography{ral.bib}

\end{document}